\documentclass[10pt,twocolumn,letterpaper]{article}
\pdfoutput=1
\usepackage{cvpr}              

\definecolor{cvprblue}{rgb}{0.21,0.49,0.74}
\usepackage[pagebackref,breaklinks,colorlinks,allcolors=cvprblue]{hyperref}
\usepackage{amsmath}
\usepackage{multirow}
\usepackage{arydshln}

\title{SymDPO: Boosting In-Context Learning of Large Multimodal Models with Symbol Demonstration Direct Preference Optimization}

\author{Hongrui Jia$^{1,\dagger}$ \quad Chaoya Jiang$^{1,\dagger}$ \quad Haiyang Xu$^{2,*}$ \quad Wei Ye$^{1,*}$ \quad Mengfan Dong$^1$ \\
Ming Yan$^2$ \quad Ji Zhang$^2$ \quad Fei Huang$^2$ \quad Shikun Zhang$^1$ \\
$^1$ National Engineering Research Center for Software Engineering, Peking University \\
$^2$ Alibaba Group \\
{\tt\small\{jiahongrui, jiangchaoya, wye, zhangsk\}@pku.edu.cn},\\
{\tt\small\{shuofeng.xhy, fei.huang\}@alibaba-inc.com}
}

\begin{document}
\maketitle
\let\thefootnote\relax\footnotetext{\noindent$^\dagger$Equal contribution. $^*$Corresponding author.}
\begin{abstract}
As language models continue to scale, Large Language Models (LLMs) have exhibited emerging capabilities in In-Context Learning (ICL), enabling them to solve language tasks by prefixing a few in-context demonstrations (ICDs) as context. Inspired by these advancements, researchers have extended these techniques to develop Large Multimodal Models (LMMs) with ICL capabilities. However, existing LMMs face a critical issue: they often fail to effectively leverage the visual context in multimodal demonstrations and instead simply follow textual patterns. This indicates that LMMs do not achieve effective alignment between multimodal demonstrations and model outputs. To address this problem, we propose Symbol Demonstration Direct Preference Optimization (SymDPO). Specifically, SymDPO aims to break the traditional paradigm of constructing multimodal demonstrations by using random symbols to replace text answers within instances. This forces the model to carefully understand the demonstration images and establish a relationship between the images and the symbols to answer questions correctly. We validate the effectiveness of this method on multiple benchmarks, demonstrating that with SymDPO, LMMs can more effectively understand the multimodal context within examples and utilize this knowledge to answer questions better. Code is available at \href{https://github.com/APiaoG/SymDPO}{https://github.com/APiaoG/SymDPO}.
\end{abstract}    
\section{Introduction}
\label{sec:intro}
\begin{figure}[htbp]
\begin{center}
\includegraphics[width=0.47\textwidth]{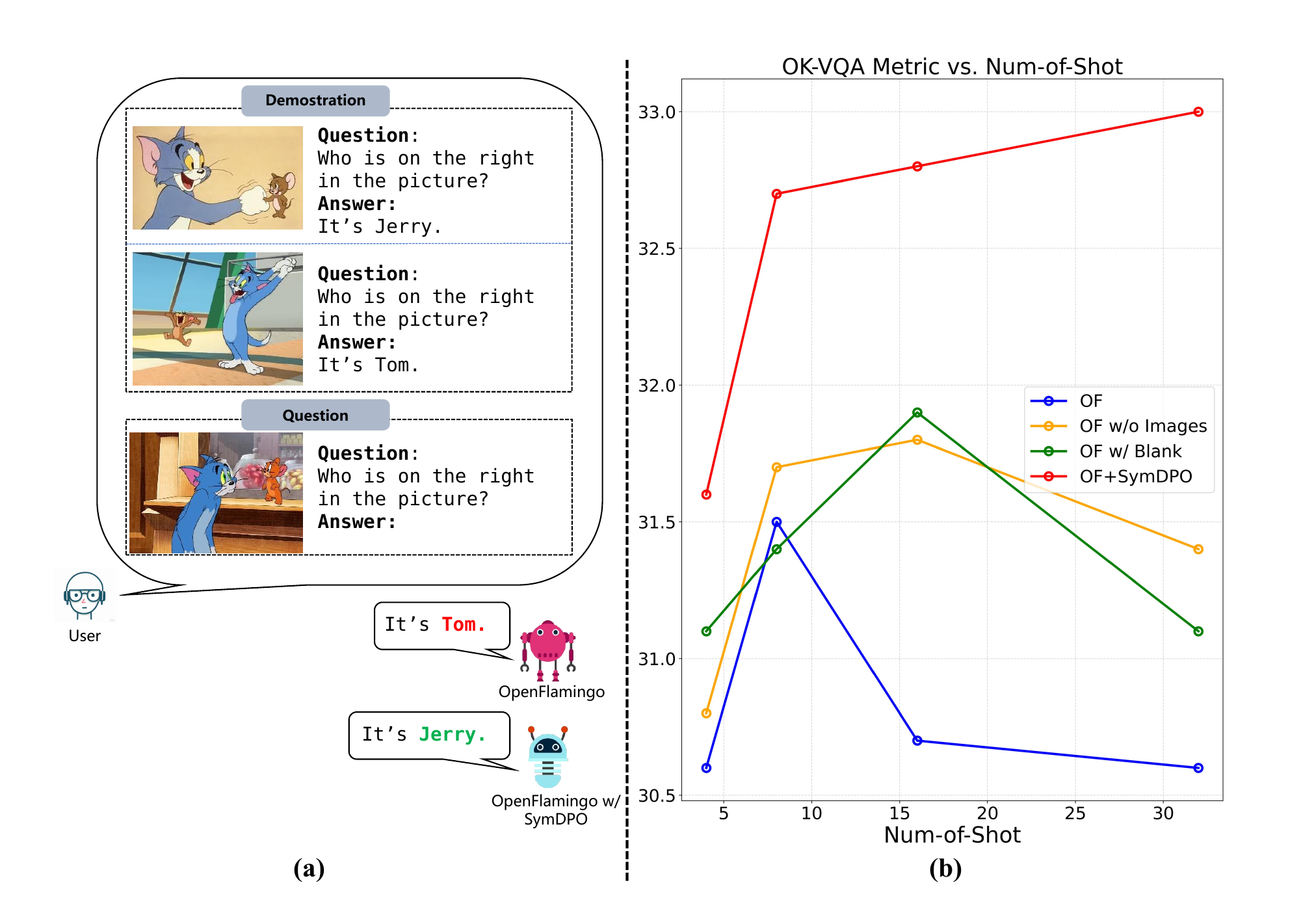}
\vspace{-3ex}
\end{center}
\caption{In subfigure (a), an example of visual context overlook is illustrated using OpenFlamingo as a case study. Here, OpenFlamingo \cite{Awadalla2023OpenFlamingoAO} erroneously generates a response by solely following the textual cues in the demonstration, leading to an inaccurate answer. After applying SymDPO to enhance alignment, OpenFlamingo with SymDPO successfully corrects its response, accurately addressing the question. Subfigure (b) further demonstrates that for OpenFlamingo (OF), replacing images in the demonstration with blank placeholders (OF w/ blank) or omitting images altogether (OF w/o image) surprisingly yields even better performance than the original setup. This result suggests a substantial model dependency on textual context over visual information.}
\label{fig:figure1}
\vspace{-4ex}
\end{figure}
The rapid advancement of Large Language Models (LLMs) \cite{brown2020language, achiam2023gpt4, ouyang2022training, touvron2023llama, bai2023qwen} has brought remarkable improvements in their In-Context Learning (ICL) capabilities \cite{dong2022ICLsurvey}. By leveraging In-Context Demonstrations (ICDs), a small set of exemplars provided as context, these models achieve impressive performance on various language tasks. This breakthrough in Natural Language Processing (NLP) has catalyzed research efforts to extend similar contextual learning capabilities to Large Multimodal Models (LMMs) \cite{2023GPT4VisionSC, alayrac2022flamingo, laurenccon2024obelics, li2023mimic,ye2023mplug,ye2024mplug,ye2024mplug3,liu2024llavaNext,liu2023llava,jiang2024hallucination,jiang2024hal}. The ultimate goal is to enable LMMs to effectively learn from a limited number of image-text pairs for specific tasks without parameter updates, thereby achieving few-shot learning in the multimodal domain.

To enhance the ICL capabilities of LMMs, prior works \cite{alayrac2022flamingo, Awadalla2023OpenFlamingoAO, laurenccon2024obelics, laurenccon2024matters,li2023mimic} have explored two primary approaches:  The first approach \cite{alayrac2022flamingo, Awadalla2023OpenFlamingoAO, Sun2023GenerativeMM,liu2024llavaNext} involves pretraining LMMs on massive-scale interleaved image-text data collected from the internet. The second approach \cite{Doveh2024TowardsMI,li2023mimic, jiang2024mantis} focuses on constructing specialized instruction-tuning datasets with numerous ICD examples. However, despite these efforts, recent studies \cite{Baldassini2024WhatMM, Chen2023UnderstandingAI} have shown that both approaches still face a significant limitation, which we term \textbf{Visual Context Overlook.} This phenomenon manifests as the LMMs persistently struggle to effectively incorporate visual context from multimodal demonstrations, exhibiting a strong bias towards textual pattern matching. As illustrated in Figure \ref{fig:figure1} (a), when presented with multimodal examples, LMMs tend to generate responses by following textual patterns in the context while failing to properly utilize the critical visual information, leading to inaccurate responses. Additionally, as shown in Figure \ref{fig:figure1} (b), substituting images in ICDs with blank images or even removing them altogether does not affect model performance, further underscoring the limited role of visual information in the current alignment process of LMMs.

This issue highlights a core limitation in LMMs' ability to follow instructions from multimodal demonstrations within ICL scenarios accurately. Recently, Direct Preference Optimization (DPO) \cite{Rafailov2023DirectPO}, a human preference-aligned reinforcement learning technique applied during the post-training phase, has been widely adopted to enhance LMMs' instruction-following capabilities \cite{liu2024llavaNext,Liu2024MIADPOMA, Pi2024StrengtheningML, Wang2024mDPOCP}, offering a promising direction for addressing visual context overlook. However, current DPO methods exhibit two key limitations within ICL scenarios: \textbf{1. Insufficient Mechanisms for Multimodal In-Context Instruction Following:} Current DPO methods \cite{Pi2024StrengtheningML, Wang2024mDPOCP} are largely optimized for general instruction-following tasks and lack the specialized mechanisms necessary to enhance LMMs’ comprehension and adherence to the combined visual and textual information characteristic of multimodal demonstrations in ICL settings. \textbf{2. Challenges in Preference Data Collection for Visual Context Dependency:} In typical Visual Question Answering (VQA) tasks, many questions can be effectively answered based on text alone, without needing information from multimodal in-context demonstrations (ICDs). This reliance on textual cues creates a significant barrier to collecting reliable preference data for multimodal learning. Specifically, it complicates the distinction between “accepted” and “rejected” answers, as models may default to simple text pattern matching.

To overcome these limitations, we introduce \textbf{SymDPO} (Symbol Demonstration Direct Preference Optimization), a novel method specifically designed to compel LMMs to depend on both visual and textual inputs in ICDs by establishing a mapping between visual information and symbolic responses. SymDPO replaces traditional textual answers in ICDs with semantically neutral or mismatched symbolic text strings—specific characters or strings that have no semantic relevance to the visual context.  This symbolic substitution compels the LMM to construct a mapping between visual elements and these symbolic strings, effectively linking the image content to a symbolic representation of the answer. As a result, the model can only generate correct responses by thoroughly interpreting the visual content within ICDs, as there is no relevant meaning in the symbolic text alone to support a response. This configuration makes visual information essential for understanding and responding accurately, ensuring that correct answers derive from a combined understanding of both image content and symbolic text. SymDPO thus redefines the model’s reliance on visual context, reinforcing the multimodal comprehension required for accurate response generation in visually-dependent scenarios. Our contributions are as follows:
\begin{itemize}
    \item  We propose a novel symbolic preference optimization method, SymDPO, that compels LMMs to leverage multimodal information effectively, advancing their ability to integrate visual and textual cues within ICDs.
    \item  We design and implement a pipeline that generates symbolic preference data, replacing textual answers with contextually mismatched symbols to enforce symbolic alignment with visual context.
    \item Through comprehensive experiments across multiple LMM architectures, we demonstrate consistent improvements in performance on various benchmarks, verifying SymDPO’s effectiveness in addressing visual context overlook and enhancing multimodal understanding.
\end{itemize}

\begin{figure*}[htbp]
\begin{center}
\includegraphics[width=\textwidth]{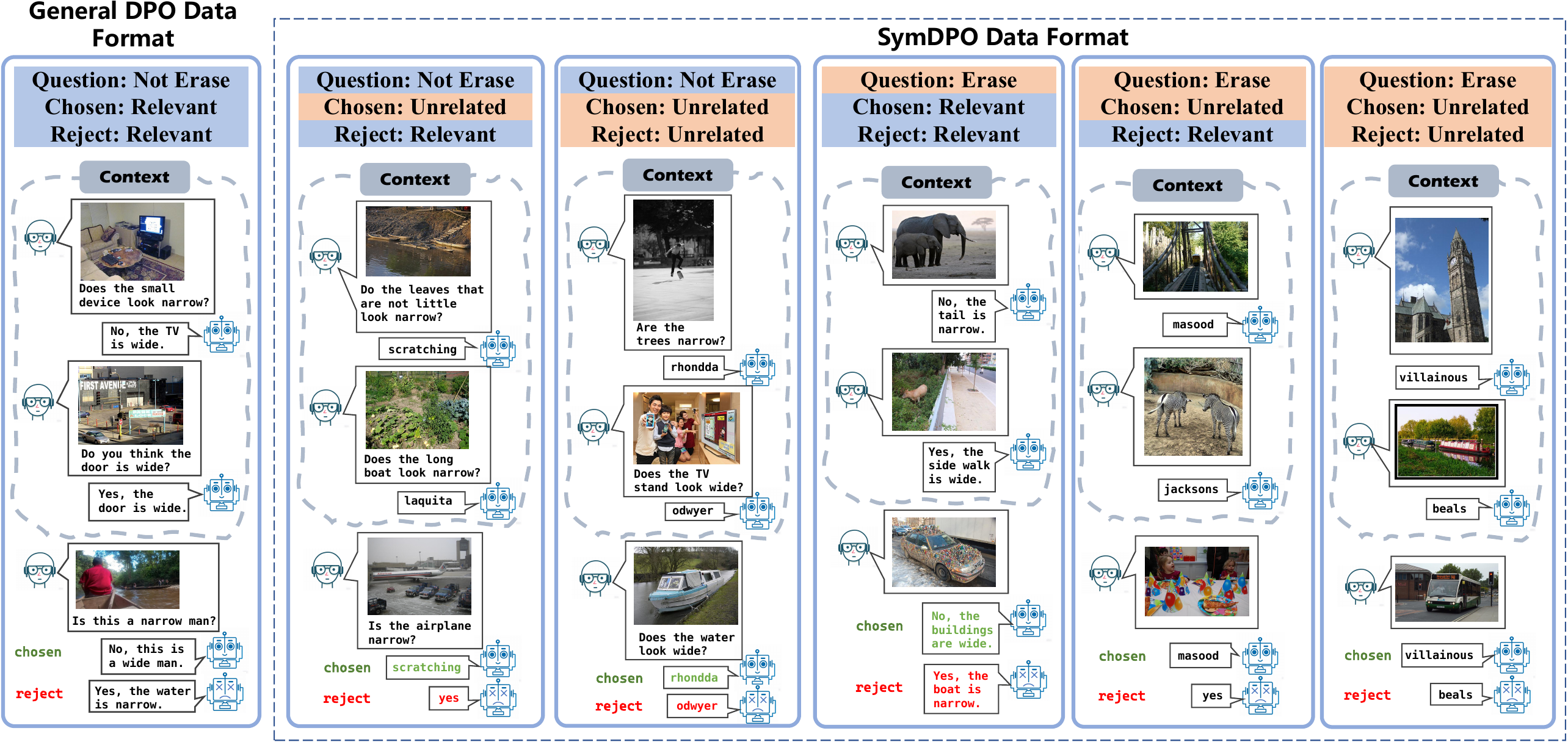}
\end{center}
\vspace{-4ex}
\caption{Comparison of General DPO and SymDPO Formats: General DPO relies solely on standard text for Questions, Answers, Chosen, and Rejected Answers, focusing on text-based training. In contrast, SymDPO replaces textual Answers with symbolized text to boost multimodal understanding, requiring models to interpret both visual and symbolized cues. This approach strengthens the model's ability to reason and decide in complex multimodal contexts.}
\label{fig:sym_struct}
\vspace{-4ex}
\end{figure*}
\section{Related Work}
\subsection{In-Context Learning}
In LLMs, prompt engineering handles specific tasks without constant fine-tuning. A variant of this method, ICL, enhances these abilities by generating prompts that incorporate multiple demonstrations \cite{brown2020language, dong2022ICLsurvey, Li2023OtterAM, Liu2022FewShotPF, Olsson2022IncontextLA}. ICL has already shown superior performance and broad applicability across many tasks \cite{min2021noisy, honovich2022instruction, mosbach2023few, hao2022structured, dong2022survey} and can be easily modified for downstream tasks \cite{Abernethy2023AMF, Wang2023HintEnhancedIL, Wies2023TheLO}. As LLMs continue to improve, more researchers are adapting them to the multimodal domain \cite{Zhao2023MMICLEV, Zhang2023MMNarratorNL, Liu2023MMHQAICLMI}. Leveraging the robust inference capabilities of LLMs, some LMMs\cite{2023GPT4VisionSC, alayrac2022flamingo, laurenccon2024obelics, li2023mimic,ye2023mplug,ye2024mplug,ye2024mplug3,liu2024llavaNext,liu2023llava,jiang2024maven} have begun to display ICL capabilities, like Flamingo \cite{alayrac2022flamingo} and IDEFICS \cite{laurenccon2024obelics}. These models have notably improved their ICL abilities by incorporating multiple samples as contextual information during the training process. However, Multimodal ICL has some limitations: LMMs pay much more attention to the textual pattern instead of image information. Previous studies \cite{yang2023lever,Zhang2023MMNarratorNL} have primarily focused on the method of constructing context to mitigate the issue, overlooking the inherent characteristics of LMMs themselves, resulting in ineffective outcomes.

\subsection{Reinforcement Learning from Human Feedback (RLHF)}
Reinforcement Learning from Human Feedback (RLHF) \cite{ouyang2022training} has become a pivotal technique in guiding model responses to align more closely with human expectations. This approach fine-tunes models based on preference data obtained from human evaluators, facilitating the development of more intuitive and contextually accurate responses \cite{Kaufmann2023ASO, Lee2023RLAIFVR}. Although RLHF has proven effective in aligning language models (LMs) with human values, its complexity and resource demands have prompted the search for alternatives. RAFT \cite{dong2023raft} chooses the best training samples through an existing reward model, while RRHF\cite{yuan2023rrhf} uses a simpler ranking loss, maintaining the efficiency of PPO \cite{PPO}. In contrast, DPO \cite{Rafailov2023DirectPO} directly optimizes LMs using a preference-based loss function, demonstrating improved training stability compared to traditional RLHF. 

\section{Method}
To address the limitations of LMMs in leveraging image information within ICL, we propose SymDPO. As shown in Figure \ref{fig:sym_struct}, SymDPO replaces the original answers in the context with unrelated words. This alteration prevents the model from depending solely on text patterns to infer answers. Instead, the model is driven to derive the correct answer by recognizing patterns that emerge jointly from the images and text in the context.  In Section 3.1, we will discuss the construction of a dataset tailored for SymDPO to facilitate this balanced learning approach. Section 3.2 will elaborate on the specific steps and mechanisms of the SymDPO algorithm.
\subsection{SymDPO Data Construction}
The construction of the SymDPO data follows a three-step process. First, we gather and structure a QA dataset in an In-Context Learning (ICL) format. Based on this ICL dataset, we proceed to build a standard DPO dataset. Finally, we introduce the concept of SymDPO, expanding upon the general DPO data foundation to create the SymDPO dataset with enhanced multimodal challenges for LMMs.

\noindent \textbf{Construct In-Context dataset:} First, we collect image-question-answer triplets from VQA datasets such as GQA, VQAv2, and image classification datasets like ImageNet. These data points are then reorganized into an In-Context Learning (ICL) format. Questions that target similar task types, such as binary yes/no questions, or those related to image object categories, attributes, relationships, or quantities, are grouped together. For example, all questions in one group may revolve around the concept of "narrow" versus "wide." Next, for each category of questions, we construct data in the ICL format:  
\( D_1, D_2, \dots, D_N, F \),  
where  
\( D_i = \{I_i, Q_i, A_i \}, i \in \{1,2,\dots, N\} \)
and  
\( F = \{\hat{I}, \hat{Q}, \hat{A}\} \).  
Here, \( D_i \) represents the \(i\)-th demonstration, and \( F \) denotes the final question-answer pair. We ensure that the demonstrations contain at least two different answers, with at least one answer matching the answer \( \hat{A} \) in \( F \). Building on this structure, we proceed to construct the DPO data required for training.

\noindent \textbf{Construct Original DPO Dataset:} For each constructed ICL dataset, we treat the original answer \( \hat{A} \) from the final QA round as the positive label. Then, we select a distinctly different answer \( A_j \) (where \( A_j \neq \hat{A} \)) from the same category of questions as the negative label, ensuring that this answer is not one of the previous answers. For instance, in the case of color-related questions, we may randomly choose another distinct answer like "The logo is black" as the negative label. With this, a single DPO data point is constructed.

\noindent \textbf{Construct SymDPO Dataset:}
The SymDPO dataset is designed to increase the difficulty of answering questions, requiring Large Multimodal Models (LMMs) to fully comprehend the combined visual and textual information within the In-Context Demonstrations (ICDs) to accurately respond. Illustrated in the Figure \ref{fig:sym_struct}, we constructed five distinct data configurations based on the standard DPO dataset to further enhance the model's comprehension capabilities.

In the SymDPO dataset, we employ a more challenging approach: firstly, all answers in the demonstrations are replaced with semantically meaningless symbols. This transformation can be expressed as:

\begin{equation}
\dot{D}_i = \{I_i, Q_i, S_i \}, \quad i \in \{1,2,\dots, N\}    
\end{equation}

where \( S_i \) represents a symbol unrelated to the actual answer, effectively stripping away semantic information to prevent the model from deducing the answer solely through simple textual patterns. This design compels the model to rely on a combination of visual and textual information within the ICDs to respond accurately within a multimodal environment.

Furthermore, a unique demonstration \( D_k \) is designated within the ICDs, in which the symbolic answer \( S_k \) aligns with the answer to the final question-answer pair \( F = \{\hat{I}, \hat{Q}, \hat{A}\} \). For this setup, the chosen answer for \( F \) is set as \( S_k \), while the rejected answer can be another unchosen answer, such as a different answer \( A_j \) from the same question type or another symbolic answer \( S_j \), provided it satisfies the following conditions:
\begin{equation}
A_j \neq \hat{A} \quad \text{and} \quad S_j \neq S_k
\end{equation}

For example, in the second data configuration of SymDPO in Figure \ref{fig:sym_struct}, we replace "narrow" and "wide" with the symbols "rhondda" and "odwyer". In this scenario, even if the model can reason independently of the ICD, it must still interpret the overall semantics of these symbols within the ICD to answer correctly. This approach ensures that the model needs to understand the implicit meaning of the symbols deeply, rather than relying solely on isolated textual or visual information.

Finally, we add further complexity by introducing additional configurations, such as whether to erase the question in the ICD or whether the Rejected Answer should be semantically relevant. These adjustments lead to five different types of SymDPO data, maintaining a certain proportion of representation across all types in the final dataset to maximize diversity.

\subsection{SymDPO}
In the previous section, we introduced the data construction process of SymDPO. In this section, we will explain the principles behind SymDPO. Training LMMs with SymDPO can make LMMs pay more attention to visual information. Aligning preferences in Large Multimodal Models (LMMs) involves aligning the model's preferences with human preferences, typically employing techniques such as RLHF\cite{ouyang2022training} (Reinforcement Learning from Human Feedback) and RLAIF\cite{bai2022constitutional} (Reinforcement Learning from AI Feedback). Considering a dataset $\mathcal {D}_{S} $, each entry includes an input $x=\{q,I,C\}$, a chosen response $y_w$ and a rejected response $ y_l $, while $q$ represents the question, $I$ represent images and $C$ represents the context. $\mathcal {D}_{S}$ can be represented as $\mathcal {D}_{S} = \{x, y_w, y_l\}$.

Upon receiving the input $x$, a LMM produces an output $ y $, to which a reward $r(x, y)$ is allocated. The reward model assesses both chosen (high $r(x, y)$) and rejected (low $r(x, y)$) samples. Meanwhile, to avoid overfitting on the dataset $\mathcal {D}_{S}$, preference alignment in LMMs incorporates a Kullback-Leibler (KL) divergence loss $D_{KL}$, which normalizes the disparity between the model's policy $\pi_\theta(y|x)$ and the reference model's policy $\pi_{ref}(y|x)$. The goal is to maximize this:


\vspace{-2ex}
\begin{equation}
\resizebox{.9\hsize}{!}{$\mathop{\max}_{\theta} {[\mathbb{E}_{x\sim\mathcal {D}_{S},(y)\sim\pi_\theta(y|x)} [r(x, y)] - \beta \cdot D_{KL}(\pi_\theta(y|x)||\pi_{ref}(y|x))]}$}
\end{equation}
Here, $\theta$ denote the parameters of the LMM, $\pi_\theta$ denote the policy of the LMM, $\pi_\theta(y|x)$ denote the distribution of the LMM and the hyperparameter $\beta$ controls the impact of the KL divergence within the optimization target. The reference model is the model's state prior to preference alignment.

To enhance the preference alignment target, the Direct Preference Optimization (DPO) method is utilized. The DPO method is efficient, stable, and does not require fitting a reward model. Our method, SymDPO, is based on the classical DPO algorithm. The SymDPO objective is formally defined as follows:

\vspace{-2ex}
\begin{small}
\begin{equation}
\begin{split}
& \mathcal {L}_{S} (\pi_\theta;\pi_{ref}) \!=\! \\
& \!-\!\mathbb{E}_{(x,y_w,y_l)\sim\mathcal {D}_{S}} log \sigma (\beta log\frac{\pi_\theta(y_w|x)}{\pi_{ref}(y_w|x)}\!-\!\beta log\frac{\pi_\theta(y_l|x)}{\pi_{ref}(y_l|x)})\!
\end{split}
\end{equation}
\end{small}


where $\sigma$ is the logistic function.

\section{Experiment}
\subsection{Experiment Setting}
\begin{table*}[t]
\setlength{\tabcolsep}{2.4mm}{
\begin{tabular}{l|c|l|c|c|c|c|c}
\toprule[1.0pt]
\scriptsize
\textbf{Model} & \textbf{Shots} & \textbf{Method} & \textbf{COCO Caption} & \textbf{Flickr-30K} & \textbf{VQAv2} & \textbf{OK-VQA} & \textbf{TextVQA} \\ 
& & & (CIDEr) & (CIDEr) & (Acc) & (Acc) & (Acc) \\ \hline

\multirow{15}{*}{OF-3B (I)} 
& \multirow{5}{*}{4} & Base & 82.7 & 59.1 & 45.7 & 30.6 & 28.1 \\ 
&  & + SymDPO & \textbf{87.4\textsubscript{\textcolor{green}{+4.7}}} & \textbf{61.2\textsubscript{\textcolor{green}{+2.1}}} & \textbf{46.2\textsubscript{\textcolor{green}{+0.5}}} &  \textbf{31.6\textsubscript{\textcolor{green}{+1.0}}} & 28.3\textsubscript{\textcolor{green}{+0.2}} \\ 
&  & + General DPO & 83.5\textsubscript{\textcolor{green}{+0.8}} & 60.0\textsubscript{\textcolor{green}{+0.9}} &  46.0\textsubscript{\textcolor{green}{+0.3}} & 30.7\textsubscript{\textcolor{green}{+0.1}} & 28.2\textsubscript{\textcolor{green}{+0.1}} \\ 
&  & + Video DPO & 82.5\textsubscript{\textcolor{red}{-0.2}} & 59.5\textsubscript{\textcolor{green}{+0.4}} & 45.5\textsubscript{\textcolor{red}{-0.2}} & 30.3\textsubscript{\textcolor{red}{-0.3}} & 28.4\textsubscript{\textcolor{green}{+0.3}} \\ 
&  & + MIA-DPO & 84.7\textsubscript{\textcolor{green}{+2.0}} & 60.8\textsubscript{\textcolor{green}{+1.7}} &  46.1\textsubscript{\textcolor{green}{+0.4}} & 30.4\textsubscript{\textcolor{red}{-0.2}} & \textbf{28.5\textsubscript{\textcolor{green}{+0.4}}} \\ 
\cline{2-8}
& \multirow{5}{*}{8} & Base & 87.8 & 60.7 & 45.9 & 31.5 & 29.1 \\ 
&  & + SymDPO & \textbf{91.2\textsubscript{\textcolor{green}{+3.4}}} & \textbf{65.3\textsubscript{\textcolor{green}{+4.6}}} & \textbf{46.5\textsubscript{\textcolor{green}{+0.6}}} & \textbf{32.7\textsubscript{\textcolor{green}{+1.2}}} & \textbf{29.8\textsubscript{\textcolor{green}{+0.7}}} \\ 
&  & + General DPO & 88.4\textsubscript{\textcolor{green}{+0.6}} & 61.5\textsubscript{\textcolor{green}{+0.8}} & 46.1\textsubscript{\textcolor{green}{+0.2}} & 31.3\textsubscript{\textcolor{red}{-0.2}} & 29.1\textsubscript{\textcolor{green}{+0.0}} \\ 
&  & + Video DPO & 87.3\textsubscript{\textcolor{red}{-0.5}} & 60.6\textsubscript{\textcolor{red}{-0.1}} & 46.0\textsubscript{\textcolor{green}{+0.1}} & 31.4\textsubscript{\textcolor{red}{-0.1}} & 28.7\textsubscript{\textcolor{red}{-0.4}} \\ 
&  & + MIA-DPO & 89.0\textsubscript{\textcolor{green}{+1.2}} & 62.5\textsubscript{\textcolor{green}{+1.8}} & 46.3\textsubscript{\textcolor{green}{+0.4}} & 31.2\textsubscript{\textcolor{red}{-0.3}} & 29.3\textsubscript{\textcolor{green}{+0.2}} \\ 
\cline{2-8}
& \multirow{5}{*}{16} & Base & 91.9 & 63.0 & 45.8 & 30.7 & 29.1 \\ 
&  & + SymDPO & \textbf{93.4\textsubscript{\textcolor{green}{+1.5}}} & \textbf{66.1\textsubscript{\textcolor{green}{+3.1}}} & \textbf{46.5\textsubscript{\textcolor{green}{+0.7}}} & \textbf{32.8\textsubscript{\textcolor{green}{+1.9}}} & \textbf{29.6\textsubscript{\textcolor{green}{+0.5}}} \\ 
&  & + General DPO & 92.0\textsubscript{\textcolor{green}{+0.1}} & 62.7\textsubscript{\textcolor{red}{-0.3}} & 46.0\textsubscript{\textcolor{green}{+0.2}} & 30.5\textsubscript{\textcolor{red}{-0.2}} & 29.0\textsubscript{\textcolor{red}{-0.1}} \\ 
&  & + Video DPO & 91.8\textsubscript{\textcolor{red}{-0.1}} & 62.8\textsubscript{\textcolor{red}{-0.2}} & 45.9\textsubscript{\textcolor{green}{+0.1}} & 30.9\textsubscript{\textcolor{green}{+0.2}} & 29.2\textsubscript{\textcolor{green}{+0.1}} \\ 
&  & + MIA-DPO & 92.5\textsubscript{\textcolor{green}{+0.6}} & 63.2\textsubscript{\textcolor{green}{+0.2}} & 46.1\textsubscript{\textcolor{green}{+0.3}} & 31.1\textsubscript{\textcolor{green}{+0.4}} & 29.4\textsubscript{\textcolor{green}{+0.3}} \\ 
\hline

\multirow{15}{*}{OF-9B} 
& \multirow{5}{*}{4} & Base & 89.0 & 65.8 & 54.8 & 40.1 & 28.2 \\ 
&  & + SymDPO & \textbf{93.8\textsubscript{\textcolor{green}{+4.8}}} & \textbf{69.4\textsubscript{\textcolor{green}{+3.6}}} & \textbf{56.8\textsubscript{\textcolor{green}{+2.0}}} & \textbf{41.0\textsubscript{\textcolor{green}{+0.9}}} & \textbf{28.8\textsubscript{\textcolor{green}{+0.6}}} \\ 
&  & + General DPO & 89.2\textsubscript{\textcolor{green}{+0.2}} & 66.4\textsubscript{\textcolor{green}{+0.6}} & 55.2\textsubscript{\textcolor{green}{+0.4}} & 40.3\textsubscript{\textcolor{green}{+0.2}} & 28.5\textsubscript{\textcolor{green}{+0.3}} \\ 
&  & + Video DPO & 88.7\textsubscript{\textcolor{red}{-0.3}} & 65.7\textsubscript{\textcolor{red}{-0.1}} & 54.7\textsubscript{\textcolor{red}{-0.1}} & 40.5\textsubscript{\textcolor{green}{+0.4}} & 28.7\textsubscript{\textcolor{green}{+0.5}} \\ 
&  & + MIA-DPO & 88.6\textsubscript{\textcolor{red}{-0.4}} & 67.5\textsubscript{\textcolor{green}{+1.7}} & 55.2\textsubscript{\textcolor{green}{+0.4}} & 40.7\textsubscript{\textcolor{green}{+0.6}} & 28.9\textsubscript{\textcolor{green}{+0.7}} \\ 
\cline{2-8}
& \multirow{5}{*}{8} & Base & 96.3 & 62.9 & 54.8 & 41.1 & 29.1 \\ 
&  & + SymDPO & \textbf{102.5\textsubscript{\textcolor{green}{+6.2}}} & \textbf{67.3\textsubscript{\textcolor{green}{+4.4}}} & \textbf{55.6\textsubscript{\textcolor{green}{+0.8}}} & \textbf{42.3\textsubscript{\textcolor{green}{+1.2}}} & \textbf{30.1\textsubscript{\textcolor{green}{+1.0}}} \\ 
&  & + General DPO & 96.5\textsubscript{\textcolor{green}{+0.2}} & 62.9\textsubscript{\textcolor{green}{+0.0}} & 55.0\textsubscript{\textcolor{green}{+0.2}} & 41.5\textsubscript{\textcolor{green}{+0.4}} & 29.3\textsubscript{\textcolor{green}{+0.2}} \\ 
&  & + Video DPO & 95.7\textsubscript{\textcolor{red}{-0.6}} & 62.8\textsubscript{\textcolor{red}{-0.1}} & 55.1\textsubscript{\textcolor{green}{+0.3}} & 40.2\textsubscript{\textcolor{red}{-0.9}} & 29.0\textsubscript{\textcolor{red}{-0.1}} \\ 
&  & + MIA-DPO & 97.0\textsubscript{\textcolor{green}{+0.7}} & 63.5\textsubscript{\textcolor{green}{+0.6}} & 55.3\textsubscript{\textcolor{green}{+0.5}} & 40.2\textsubscript{\textcolor{red}{-0.9}} & 29.7\textsubscript{\textcolor{green}{+0.6}} \\ 
\cline{2-8}
& \multirow{5}{*}{16} & Base & 98.8 & 62.8 & 54.3 & 42.7 & 27.3 \\ 
&  & + SymDPO & \textbf{104.3\textsubscript{\textcolor{green}{+5.5}}} & \textbf{64.9\textsubscript{\textcolor{green}{+2.1}}} & \textbf{55.7\textsubscript{\textcolor{green}{+1.4}}} & \textbf{44.5\textsubscript{\textcolor{green}{+1.8}}} & \textbf{28.1\textsubscript{\textcolor{green}{+0.8}}} \\ 
&  & + General DPO & 98.9\textsubscript{\textcolor{green}{+0.1}} & 63.0\textsubscript{\textcolor{green}{+0.2}} & 54.5\textsubscript{\textcolor{green}{+0.2}} & 41.9\textsubscript{\textcolor{red}{-0.8}} & 27.5\textsubscript{\textcolor{green}{+0.2}} \\ 
&  & + Video DPO & 98.2\textsubscript{\textcolor{red}{-0.6}} & 62.2\textsubscript{\textcolor{red}{-0.6}} & 54.6\textsubscript{\textcolor{green}{+0.3}} & 42.7\textsubscript{\textcolor{green}{+0.0}} & 26.7\textsubscript{\textcolor{red}{-0.6}} \\ 
&  & + MIA-DPO & 98.5\textsubscript{\textcolor{red}{-0.3}} & 62.9\textsubscript{\textcolor{green}{+0.1}} & 54.8\textsubscript{\textcolor{green}{+0.5}} & 43.1\textsubscript{\textcolor{green}{+0.4}} & 26.9\textsubscript{\textcolor{red}{-0.4}} \\ 
\hline

\multirow{15}{*}{IDEFICS-9B} 
& \multirow{5}{*}{4} & Base & 93.0 & 59.7 & 55.4 & 45.4 & 27.6 \\ 
&  & + SymDPO & \textbf{96.5\textsubscript{\textcolor{green}{+3.5}}} & \textbf{64.0}\textsubscript{\textcolor{green}{+4.3}} & \textbf{56.1\textsubscript{\textcolor{green}{+0.7}}} & \textbf{47.2\textsubscript{\textcolor{green}{+1.8}}} & \textbf{28.6\textsubscript{\textcolor{green}{+1.0}}} \\ 
&  & + General DPO & 93.2\textsubscript{\textcolor{green}{+0.2}} & 60.2\textsubscript{\textcolor{green}{+0.5}} & 55.6\textsubscript{\textcolor{green}{+0.2}} & 45.9\textsubscript{\textcolor{green}{+0.5}} & 27.8\textsubscript{\textcolor{green}{+0.2}} \\ 
&  & + Video DPO & 93.5\textsubscript{\textcolor{green}{+0.5}} & 59.6\textsubscript{\textcolor{red}{-0.1}} & 55.7\textsubscript{\textcolor{green}{+0.3}} & 45.8\textsubscript{\textcolor{green}{+0.4}} & 28.0\textsubscript{\textcolor{green}{+0.4}} \\ 
&  & + MIA-DPO & 93.7\textsubscript{\textcolor{green}{+0.7}} & 61.5\textsubscript{\textcolor{green}{+1.8}} & 55.9\textsubscript{\textcolor{green}{+0.5}} & 46.3\textsubscript{\textcolor{green}{+0.9}} & 28.2\textsubscript{\textcolor{green}{+0.6}} \\ 
\cline{2-8}
& \multirow{5}{*}{8} & Base & 97.0 & 61.9 & 56.4 & 47.7 & 27.5 \\ 
&  & + SymDPO & \textbf{103.8\textsubscript{\textcolor{green}{+6.8}}} & \textbf{66.1\textsubscript{\textcolor{green}{+4.2}}} & \textbf{57.2\textsubscript{\textcolor{green}{+0.8}}} & \textbf{49.5\textsubscript{\textcolor{green}{+1.8}}} & \textbf{28.5\textsubscript{\textcolor{green}{+1.0}}} \\ 
&  & + General DPO & 97.2\textsubscript{\textcolor{green}{+0.2}} & 62.0\textsubscript{\textcolor{green}{+0.1}} & 56.6\textsubscript{\textcolor{green}{+0.2}} & 48.1\textsubscript{\textcolor{green}{+0.4}} & 27.7\textsubscript{\textcolor{green}{+0.2}} \\ 
&  & + Video DPO & 97.5\textsubscript{\textcolor{green}{+0.5}} & 62.2\textsubscript{\textcolor{green}{+0.3}} & 56.2\textsubscript{\textcolor{red}{-0.2}} & 47.3\textsubscript{\textcolor{red}{-0.4}} & 27.9\textsubscript{\textcolor{green}{+0.4}} \\ 
&  & + MIA-DPO & 97.7\textsubscript{\textcolor{green}{+0.7}} & 62.5\textsubscript{\textcolor{green}{+0.6}} & 56.9\textsubscript{\textcolor{green}{+0.5}} & 48.3\textsubscript{\textcolor{green}{+0.6}} & 28.1\textsubscript{\textcolor{green}{+0.6}} \\ 
\cline{2-8}
& \multirow{5}{*}{16} & Base & 99.7 & 64.5 & 57.0 & 48.4 & 27.9 \\ 
&  & + SymDPO & \textbf{107.9\textsubscript{\textcolor{green}{+8.2}}} & \textbf{69.3\textsubscript{\textcolor{green}{+4.8}}} & \textbf{58.2\textsubscript{\textcolor{green}{+1.2}}} & \textbf{50.6\textsubscript{\textcolor{green}{+2.2}}} & \textbf{29.3\textsubscript{\textcolor{green}{+1.4}}} \\ 
&  & + General DPO & 99.6\textsubscript{\textcolor{red}{-0.1}} & 64.7\textsubscript{\textcolor{green}{+0.2}} & 57.2\textsubscript{\textcolor{green}{+0.2}} & 43.8\textsubscript{\textcolor{red}{-4.6}} & 28.8\textsubscript{\textcolor{green}{+0.9}} \\ 
&  & + Video DPO & 99.4\textsubscript{\textcolor{red}{-0.3}} & 63.9\textsubscript{\textcolor{red}{-0.6}} & 57.3\textsubscript{\textcolor{green}{+0.3}} & 48.3\textsubscript{\textcolor{red}{-0.1}} & 28.0\textsubscript{\textcolor{green}{+0.1}} \\ 
&  & + MIA-DPO & 99.8\textsubscript{\textcolor{green}{+0.1}} & 64.0\textsubscript{\textcolor{red}{-0.5}} & 57.5\textsubscript{\textcolor{green}{+0.5}} & 48.2\textsubscript{\textcolor{red}{-0.2}} & 28.2\textsubscript{\textcolor{green}{+0.3}} \\ 
\hline

\bottomrule
\end{tabular}}
\centering
\caption{Comparison of Different DPO Methods: Performance of models with various DPO techniques across benchmarks and shot counts.}
\vspace{-4ex}
\label{table:main}
\end{table*}
\begin{figure*}[t]
\begin{center}
\includegraphics[width=\textwidth]{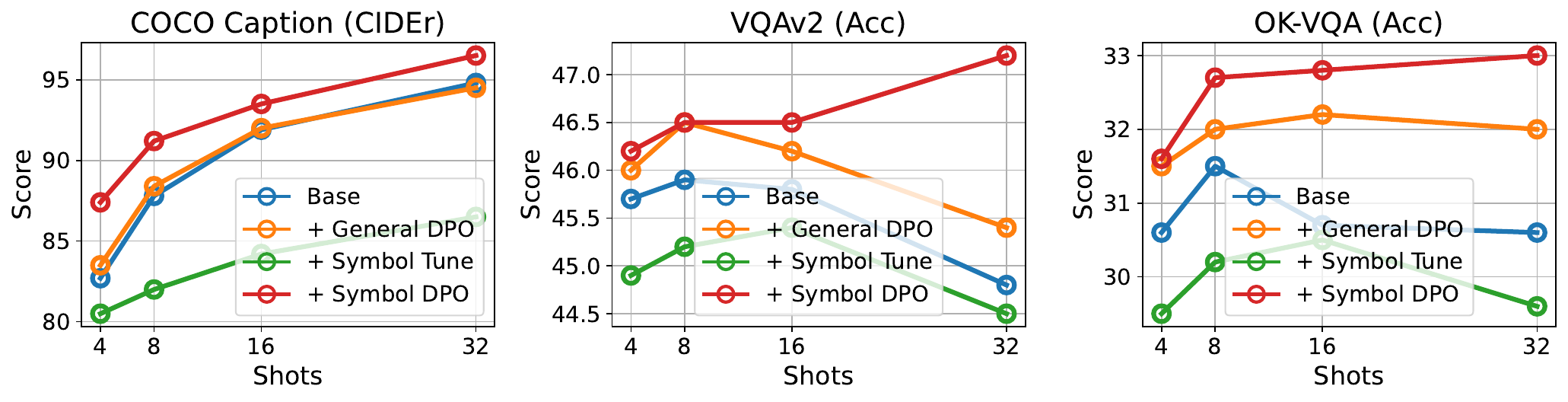}
\end{center}
\vspace{-5ex}
\caption{Comparison of Symbol Tuning, General DPO, and SymDPO Methods: We optimized OF 3b using three different methods: Symbol Tuning, General DPO, and SymDPO, resulting in three distinct variants. The performance of these variants was visualized using line charts, showcasing the results across four-shot (4, 8, 16, 32) settings on the COCO, VQAv2, and OK-VQA benchmarks.}
\label{fig:sym_tune}
\vspace{-3ex}
\end{figure*}

\noindent \textbf{Implementation Details:} We instantiate our model with Open-Flamingo \cite{Awadalla2023OpenFlamingoAO} and IDEFICS \cite{Pi2024StrengtheningML}, constructing the SymDPO dataset from the VQAv2 \cite{antol2015vqa}, GQA \cite{hudson2019gqa}, and ImageNet \cite{deng2009imagenet} training sets, amassing a total of 872,000 data items. From this collection, a subset of 10,000 samples is randomly selected for training. To enhance data quality, we apply GPT-4v to the selected samples. In the post-training phase, we employ linear annealing to adjust the learning rate, initializing it at 5e-6. This task is executed on 8 NVIDIA A100 GPUs, with the complete post-training process taking approximately 1 hour.

\noindent \textbf{Benchmark:} Our model is evaluated in alignment with Flamingo on image captioning benchmarks, speci  fically COCO Caption \cite{chen2015microsoft} and Flickr 30K \cite{young2014image}, as well as on three question-answering (QA) benchmarks: VQA v2 \cite{antol2015vqa}, OK-VQA \cite{marino2019ok}, and TextVQA \cite{singh2019towards}. For the image captioning tasks, we report CIDEr scores \cite{vedantam2015cider} as evaluation metrics, while for QA tasks, we use accuracy (Acc) as the metric.

\noindent \textbf{Baseline:}
We compare SymDPO against two different DPO optimization approaches: 
\begin{itemize}
    \item Video DPO \cite{zhang2024direct} - Proposed by LLaVA-Hound-DPO, this approach utilizes a video-specific DPO dataset to improve the model’s understanding of video data.
    \item MIA-DPO \cite{liu2024mia} - Designed for multi-image scenarios, this method aims to mitigate hallucinations in LMMs by optimizing in multi-image settings.
\end{itemize}

\subsection{Main Results}
As illustrated in Table \ref{table:main}, we evaluated the performance of SymDPO, Video DPO, and MIA-DPO on Open-Flamingo (OF) and IDEFICS 9B across five different benchmarks. The results reveal that SymDPO consistently enhances performance across all benchmarks for both OF and IDEFICS, demonstrating the efficacy of SymDPO. In contrast, Video DPO showed no notable improvement, while MIA-DPO yielded only marginal gains. We attribute these outcomes to the specific design focuses of Video DPO and MIA-DPO: Video DPO is primarily oriented toward semantic alignment and optimization for video data, whereas MIA-DPO targets alignment for generic multi-image instructions. Neither approach explicitly addresses the instruction alignment in in-context scenarios, a key focus of SymDPO. We interpret this as a result of the expanded contextual knowledge composed of both visual and textual elements, allowing an LMM fine-tuned with SymDPO to better integrate and leverage this combined knowledge, thus achieving greater performance gains.

\subsection{Ablation Study}
\subsubsection{Effectiveness of SymDPO}



\begin{figure*}[t]
\begin{center}
\includegraphics[width=\textwidth]{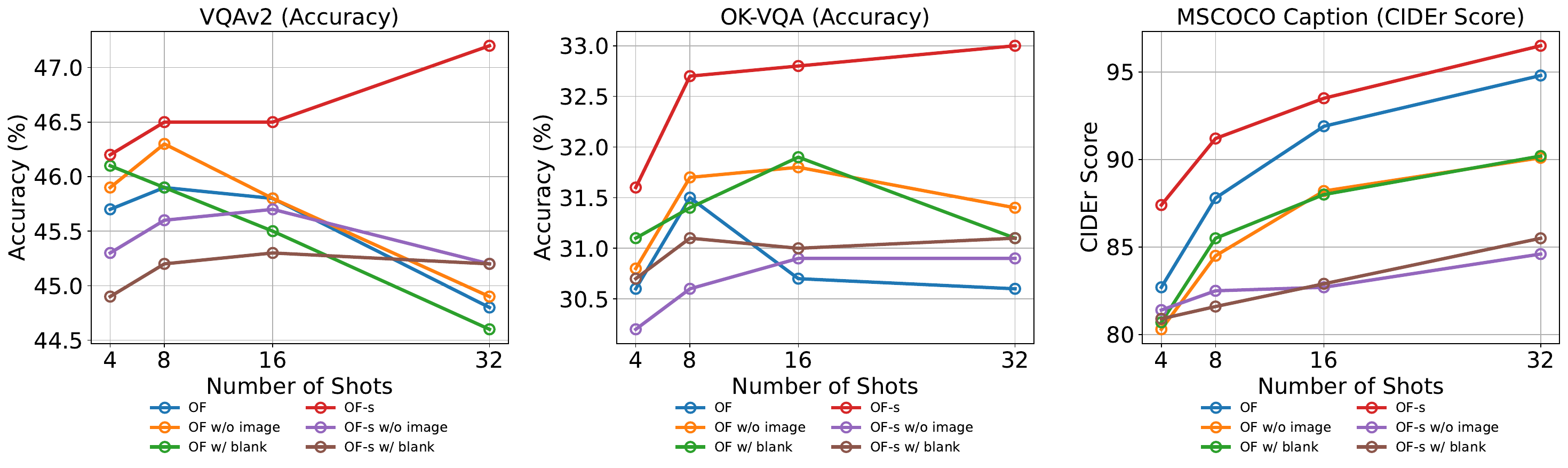}
\end{center}
\vspace{-3ex}
\caption{Impact of Visual Context Removal on OF and OF+SymDPO Performance.}
\label{fig:effect_noimage}
\vspace{-1ex}
\end{figure*}




\begin{table*}[t]
\setlength{\tabcolsep}{2.0mm}{
\begin{tabular}{l|c|l|c|c|c|c|c}
\toprule[1.0pt]
\small
\textbf{Model} & \textbf{Shots} & \textbf{Method} & \textbf{COCO Caption} & \textbf{Flickr-30K} & \textbf{VQAv2} & \textbf{OK-VQA} & \textbf{TextVQA} \\ 
& & & (CIDEr) & (CIDEr) & (Acc) & (Acc) & (Acc) \\ \hline

\multirow{9}{*}{OF-3B (I)} 
& \multirow{4}{*}{4} & Base & 82.7 & 59.1 & 45.7 & 30.6 & 28.1 \\ 
&  & + RICES  & 90.5$_{\textcolor{green}{+7.8}}$ & 53.9$_{\textcolor{red}{-5.2}}$ & 45.3$_{\textcolor{red}{-0.4}}$ & 31.4$_{\textcolor{green}{+0.8}}$ &  28.9$_{\textcolor{green}{+0.8}}$ \\
&  & + SymDPO & 87.4$_{\textcolor{green}{+4.7}}$ & 61.2$_{\textcolor{green}{+2.1}}$ & 45.8$_{\textcolor{green}{+0.1}}$ & 31.6$_{\textcolor{green}{+1.0}}$ & 28.3$_{\textcolor{green}{+0.2}}$ \\ 
&  & + SymDPO \& RICES & \textbf{93.5$_{\textcolor{green}{+10.8}}$} & \textbf{62.0$_{\textcolor{green}{+2.9}}$} & \textbf{46.6$_{\textcolor{green}{+0.9}}$} & \textbf{33.4$_{\textcolor{green}{+2.8}}$} & \textbf{29.1$_{\textcolor{green}{+1.0}}$} \\ 
\cline{2-8}
& \multirow{4}{*}{8} & Base & 87.8 & 60.7 & 45.9 & 31.5 & 29.1 \\ 
&  & + RICES & 96.8$_{\textcolor{green}{+9.0}}$ & 58.6$_{\textcolor{red}{-2.1}}$ & 46.1$_{\textcolor{green}{+0.2}}$ & 32.8$_{\textcolor{green}{+1.3}}$ & 28.8$_{\textcolor{red}{-0.3}}$ \\
&  & + SymDPO & 91.2$_{\textcolor{green}{+3.4}}$ & 65.3$_{\textcolor{green}{+4.6}}$ & 46.5$_{\textcolor{green}{+0.6}}$ & 32.7$_{\textcolor{green}{+1.2}}$ & 29.8$_{\textcolor{green}{+0.7}}$ \\
&  & + SymDPO \& RICES & \textbf{98.4$_{\textcolor{green}{+10.6}}$} & \textbf{68.2$_{\textcolor{green}{+7.5}}$} & \textbf{47.2$_{\textcolor{green}{+1.3}}$} & \textbf{34.3$_{\textcolor{green}{+2.8}}$} & \textbf{31.7$_{\textcolor{green}{+2.6}}$} \\ 
\cline{2-8}
& \multirow{4}{*}{16} & Base & 91.9 & 63.0 & 45.8 & 30.7 & 29.1 \\ 
&  & + RICES & 101.1$_{\textcolor{green}{+9.2}}$ & 61.5$_{\textcolor{red}{-1.5}}$ & 46.6$_{\textcolor{green}{+0.8}}$ & 33.9$_{\textcolor{green}{+3.2}}$ & 28.8$_{\textcolor{red}{-0.3}}$ \\
&  & + SymDPO & 93.4$_{\textcolor{green}{+1.5}}$ & 64.6$_{\textcolor{green}{+1.6}}$ & 46.5$_{\textcolor{green}{+0.7}}$ & 32.8$_{\textcolor{green}{+1.9}}$ & 29.6$_{\textcolor{green}{+0.5}}$ \\ 
&  & + SymDPO \& RICES & \textbf{106.8$_{\textcolor{green}{+14.9}}$} & \textbf{66.5$_{\textcolor{green}{+3.5}}$} & \textbf{47.2$_{\textcolor{green}{+1.4}}$} & \textbf{35.1$_{\textcolor{green}{+4.4}}$} & \textbf{29.8$_{\textcolor{green}{+0.7}}$} \\ 
\bottomrule
\end{tabular}}
\centering
\vspace{-1ex}
\caption{Performance comparison of the OF-3B (I) model using RICES and SymDPO across various datasets and shot counts.}
\vspace{-3ex}
\label{table:effect_RICES}
\end{table*}

To further assess the effectiveness of SymDPO, we conducted several ablation experiments.

\noindent \textbf{General DPO vs. SymDPO}: As outlined in Method Subsection 3.1, the General DPO approach employs a standard DPO dataset for optimization, without replacing answers with symbols as SymDPO does. As shown in Table \ref{table:main} and Figure \ref{fig:sym_tune}, we observed that models optimized with General DPO (i.e., Open-Flamingo (OF) and IDEFICS) exhibit significantly lower performance improvements compared to those optimized with SymDPO. This result substantiates the advantage of the symbolic answer replacement strategy within SymDPO, affirming its effectiveness.

\noindent \textbf{Visual Context Overlook Investigation}:  
   To determine whether SymDPO’s enhancements arise from addressing the visual context overlook issue in large multimodal models, we conducted additional tests. Specifically, we modified the demonstration data by either replacing images with blank placeholders ("w/ blank") or omitting images altogether ("w/o image"). We then evaluated the performance of OF and OF+SymDPO ("OF-s") under these modified conditions. The results, displayed in Figure \ref{fig:effect_noimage}, reveal that the performance of OF-SymDPO significantly declines when images are removed, suggesting that the model’s advantage derives from its comprehensive understanding of both visual and textual information in in-context demonstrations, rather than relying solely on textual data. This further emphasizes SymDPO's capability in leveraging the integrated visual-textual knowledge, enabling a more robust and contextually aware model.

\begin{figure*}[htbp]
\begin{center}
\includegraphics[width=\textwidth]{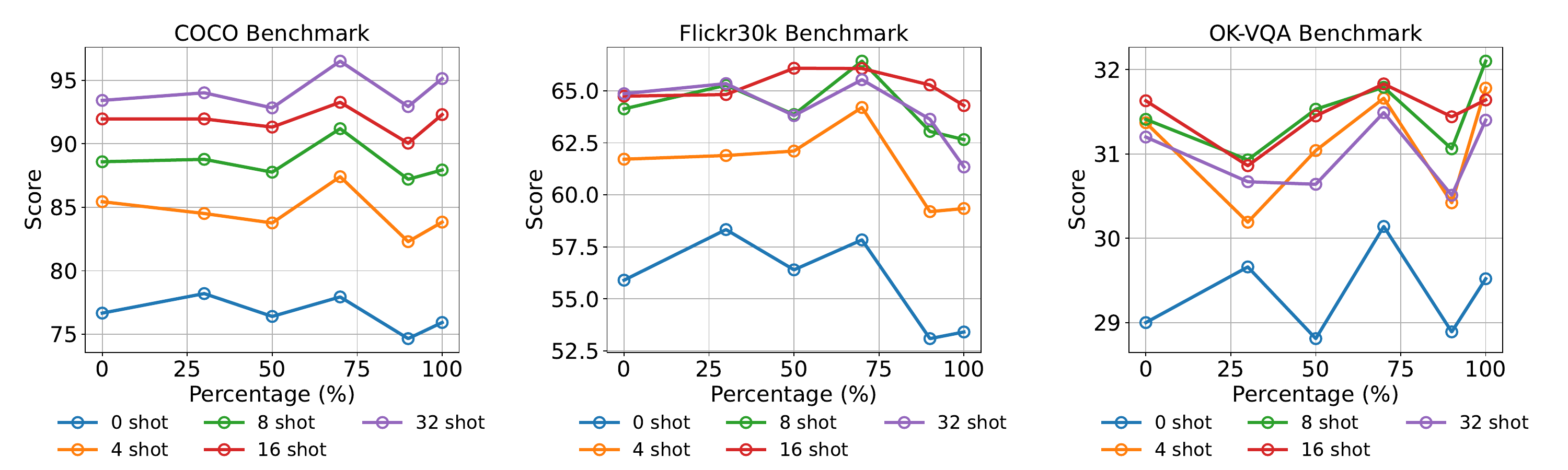}
\end{center}
\vspace{-2ex}
\caption{Comparison of the Impact of General DPO and SymDPO on LMMs with Varying Data Proportions in the Preference Dataset}
\label{fig:effect_sym}
\end{figure*}
\begin{figure*}[t]
\begin{center}
\includegraphics[width=0.98\textwidth]{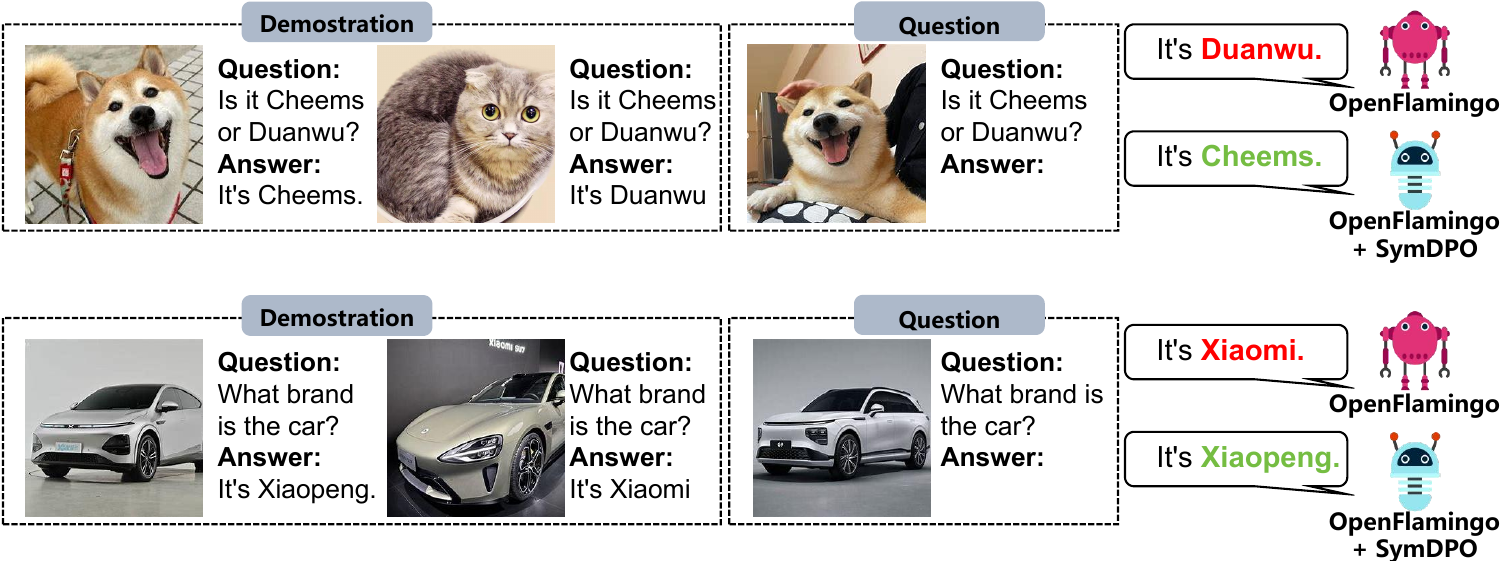}
\end{center}
\vspace{-3ex}
\caption{Example Visualization of OpenFlamingo 3B and OpenFlamingo 3B + SymDPO in ICL 2-Shot Setting.}
\label{fig:case-study}
\vspace{-1ex}
\end{figure*}
\noindent \textbf{Symbol DPO VS. Symbol Finetuning:}
To validate the advantages of DPO-based optimization, we conducted additional experiments following the preference data collection methodology in Method Subsection 3.1. After collecting the preference dataset, we constructed a multimodal symbolic fine-tuning dataset, inspired by the Symbol Tuning approach \cite{wei2023symbol}. In this setup, we used the chosen answer as the target label for an autoregressive generation task during model fine-tuning, producing the variant OF3B + SymTune. The experimental results, as shown in Table \ref{fig:sym_tune}, indicate that SymTune does not achieve satisfactory outcomes; notably, its performance on captioning tasks even declines. In contrast, the performance gains of OF3B + SymDPO are substantial across all benchmarks. We attribute this difference to the following key factors: The SymTune approach relies on symbolic fine-tuning where the model learns to predict the chosen answer directly in an autoregressive manner. However, this approach may not fully exploit preference data’s structured feedback, resulting in limited guidance for multimodal alignment.

\subsubsection{Effect of Different Demonstration Selection Strategies}
In the ICL scenario, the choice of demonstration examples (demos) can significantly influence the reasoning performance of large multimodal models. However, prior research has also noted that, due to the issue of visual context overlooks, varying the demo selection does not markedly impact LMM performance. To validate the effectiveness of SymDPO, we employed the RICES (Retrieval In-Context Example Selection) method, as used in Flamingo, to select demos. We then re-evaluated the performance of OpenFlamingo 3B (OF) and OF + SymDPO across different benchmarks. As shown in Table \ref{table:effect_RICES}, introducing RICES leads to a more pronounced improvement in the SymDPO model. This finding further highlights that incorporating SymDPO enables LMMs to leverage the integrated semantics and knowledge present in the demo’s visual-textual content more effectively.

\subsubsection{SymDPO and General DPO Integration: Impact on Task Performance}

As depicted in Figure \ref{fig:effect_sym}, we investigated the effects of integrating SymDPO and General DPO at varying proportions on task-specific performance throughout the alignment phase of DPO optimization. Symbolic data was incrementally introduced into the OF optimization process at ratios of 0\%, 30\%, 50\%, 70\%, and 100\%, and model performance was subsequently evaluated across benchmark datasets. The experimental results presented in Figure \ref{fig:effect_sym} indicate that model performance improves with increasing proportions of symbolic data, particularly within question-answering (QA) tasks. However, an exclusive reliance on SymDPO data does not yield optimal performance. Our findings show that a 70\% symbolic data ratio achieves peak effectiveness; whereas a 100\% symbolic data ratio is more effective for the OK-VQA task, suggesting task-specific dependencies on the symbolic data ratio.

\subsection{Case Study}
For further quantitative analysis of our method’s effectiveness, we visualized several diverse In-Context Learning (ICL) scenarios. As illustrated in Figure \ref{fig:case-study}, OF+SymDPO consistently yields accurate answers by interpreting the semantic context within demonstrations, whereas OF alone often misinterprets the task, relying predominantly on proximate textual information rather than fully understanding the demonstration content. In the first case, a demonstration provides the names of two pets: a dog named "Cheems" and a cat named "Duanwu." When the model is shown an image of the dog and asked to identify its name, OF+SymDPO accurately answers "Cheems," whereas OF responds incorrectly with "Duanwu," influenced by nearby textual cues without integrating the visual context. This pattern recurs across other cases, indicating that SymDPO effectively addresses the "visual context overlook" issue in Large Multimodal Models (LMMs). By doing so, SymDPO enables these models to comprehend and utilize both visual and textual information from demonstrations in a more holistic manner.




\section{Conclusion}
This work presented SymDPO, a symbolic preference optimization method designed to tackle the visual context overlooked in LMMs. By enforcing reliance on both visual and textual cues in In-Context Demonstrations, SymDPO effectively reduces LMMs' tendency toward textual pattern matching. Experiments confirm that SymDPO improves multimodal comprehension by compelling models to integrate visual context meaningfully, leading to consistent performance gains across benchmarks. In sum, SymDPO provides a robust approach to enhancing multimodal learning, marking a step toward more contextually aware LMMs.
{
    \small
    \bibliographystyle{ieeenat_fullname}
    \bibliography{main}
}


\end{document}